%% file: main.tex
\documentclass[twocolumn,letterpaper]{IEEEAerospaceCLS}
\usepackage[]{graphicx,float,latexsym,amssymb,amsfonts,amsmath,amstext,times,url}
\usepackage{enumitem}
\setlist[itemize]{noitemsep, topsep=0pt}
\usepackage[frozencache,cachedir=.]{minted} 
\usepackage{tcolorbox}
\usepackage{xcolor}
\usepackage{subcaption}
\usepackage[inkscapelatex=false]{svg}

\setlist[itemize]{itemsep=3pt}
\setlist[enumerate]{itemsep=3pt}

\definecolor{thoughtColor}{HTML}{B8E986}
\definecolor{actColor}{HTML}{FFF4E2}
\definecolor{obsColor}{HTML}{E2F0FB}
\definecolor{background}{HTML}{F8F8F8}

\newcommand{\ignore}[1]{}

\makeatletter
\renewcommand\subsection{\@startsection{subsection}{2}{\z@}%
  {-3.25ex \@plus -1ex \@minus -.2ex}%
  {1.5ex \@plus .2ex}%
  {\normalfont\bfseries}}
\makeatother

\tcbset{
  mybox/.style={
    colback=white,
    colframe=gray!50,
    sharp corners,
    boxrule=0.5mm,
    left=2mm, right=2mm, top=1mm, bottom=1mm,
    width=\linewidth,
    enlarge left by=0mm,
    enlarge right by=0mm,
    fonttitle=\bfseries,
  },
  thought/.style={colback=thoughtColor, colframe=gray!80, fonttitle=\bfseries\footnotesize},
  act/.style={colback=actColor, colframe=gray!80, fonttitle=\bfseries\footnotesize},
  obs/.style={colback=obsColor, colframe=gray!80, fonttitle=\bfseries\footnotesize},
}

\newtcolorbox{interactionbox}{
  colback=gray!5,
  colframe=gray!80,
  boxrule=0.5pt,
  arc=3pt,
  left=5pt,
  right=5pt,
  top=5pt,
  bottom=5pt
}

\usepackage{color}
\usepackage{flushend}
\usepackage{url}
\usepackage[breaklinks]{hyperref}
\usepackage[capitalise]{cleveref}
\usepackage{booktabs}
\usepackage{makecell}
\usepackage{multirow}

\begin{document}

\title{Enabling Novel Mission Operations and Interactions with ROSA: The Robot Operating System Agent}

\author{%
Rob Royce, Marcel Kaufmann, Jonathan Becktor, Sangwoo Moon, Kalind Carpenter,\\
Kai Pak, Amanda Towler, Rohan Thakker, Shehryar Khattak\\
\\
NASA Jet Propulsion Laboratory\\
California Institute of Technology\\
4800 Oak Grove Dr\\
Pasadena, CA 91109\\
\\
\{rob.royce; marcel.kaufmann; jonathan.becktor; sangwoo.moon; kalind.carpenter; \\
kai.pak; amanda.towler;
rohan.a.thakker; 
skhattak\}@jpl.nasa.gov\\
\thanks{\footnotesize \copyright2025 IEEE. Personal use of this pre-print is permitted. Permission from IEEE must be obtained for all other uses.}
}

\maketitle
\thispagestyle{plain}
\pagestyle{plain}

\begin{figure*}[htbp]
    \centering
    \begin{subfigure}{0.33\textwidth}
        \centering
        \includegraphics[width=\linewidth,height=3.75cm]{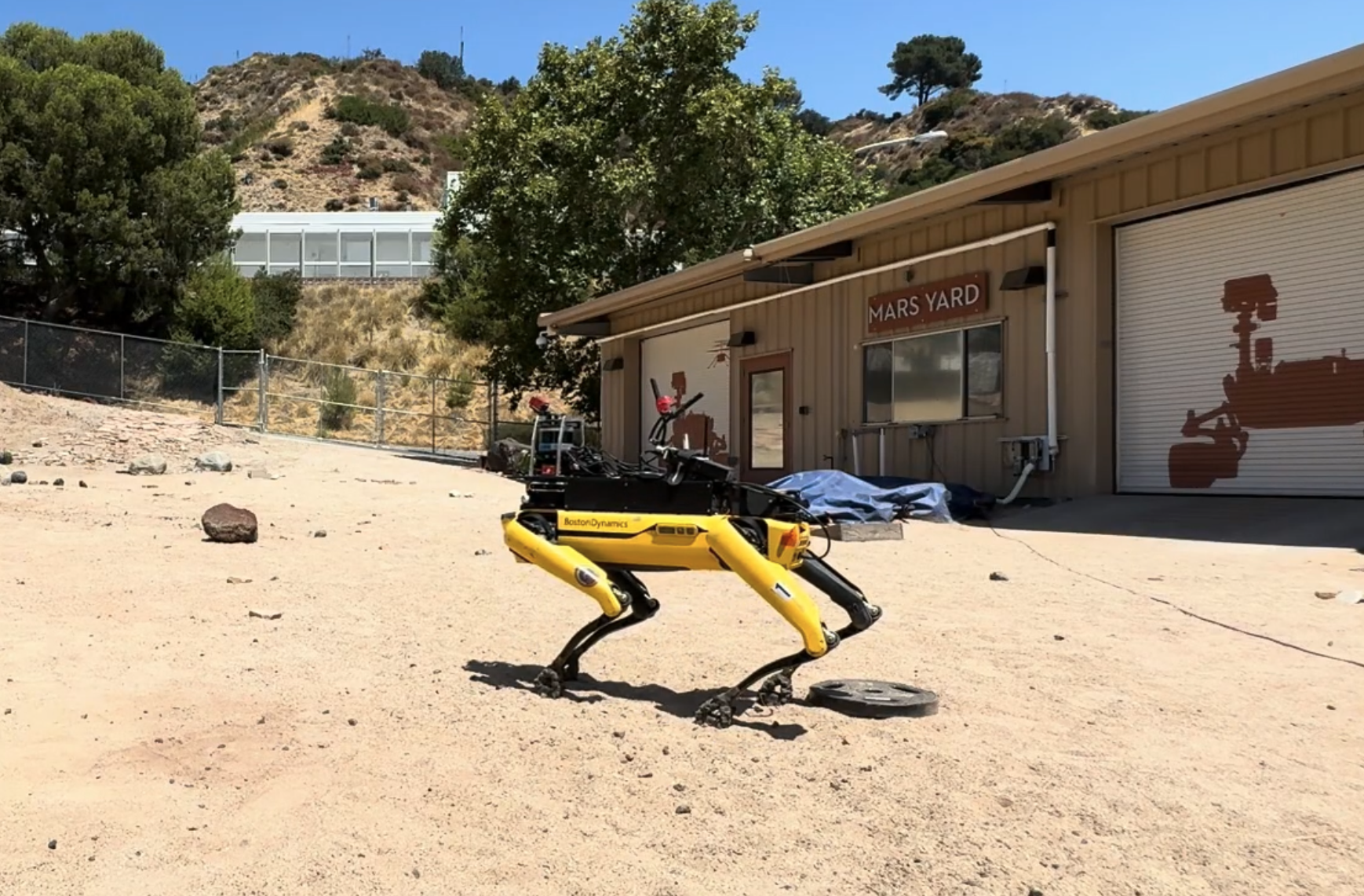}
        \caption{NeBula-Spot in the Mars Yard}
    \end{subfigure}
    \hfill
    \begin{subfigure}{0.33\textwidth}
        \centering
        \includegraphics[width=\linewidth,height=3.75cm]{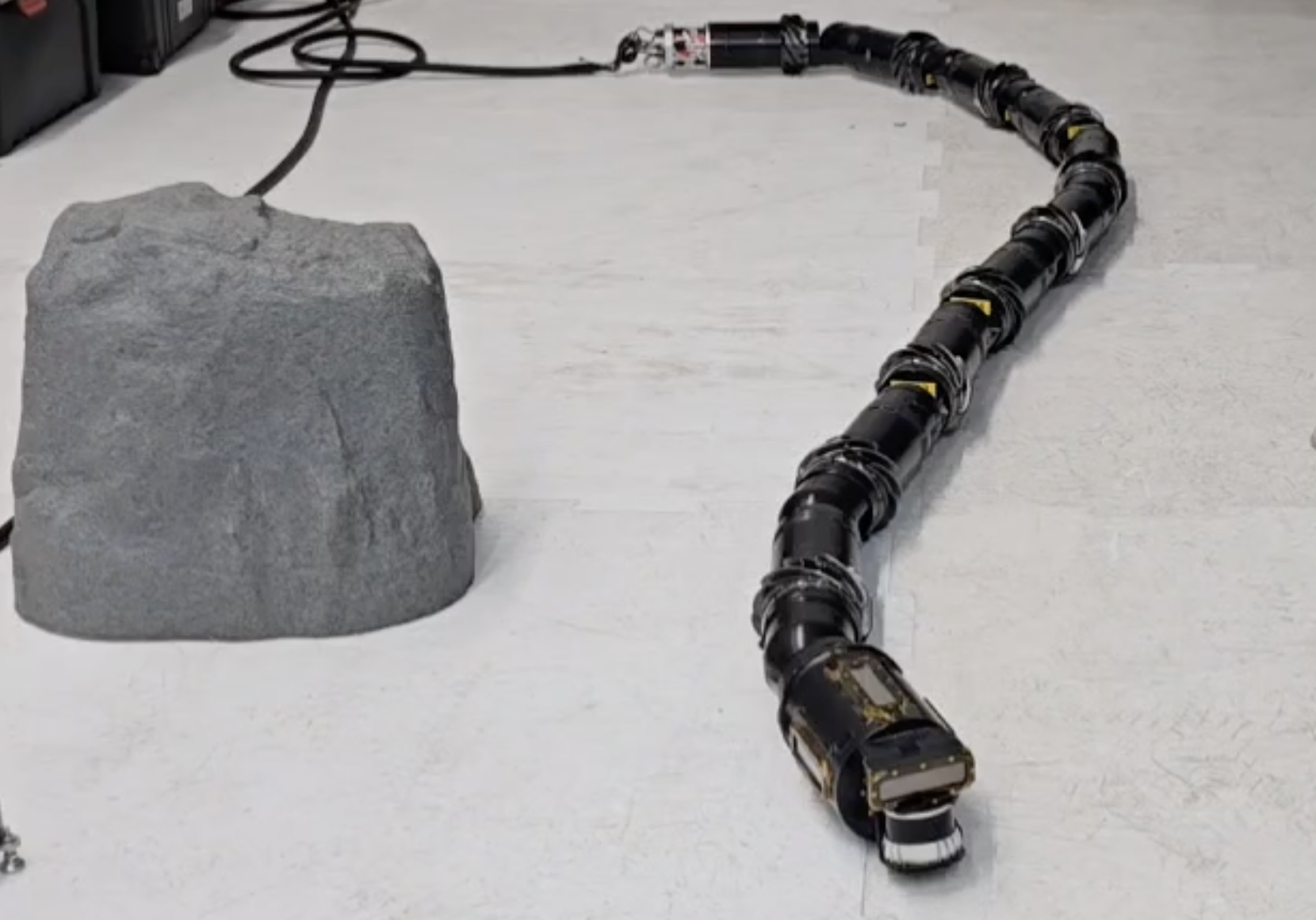}
        \caption{EELS in Laboratory Environment}
    \end{subfigure}
    \hfill
    \begin{subfigure}{0.33\textwidth}
        \centering
        \includegraphics[width=\linewidth,height=3.75cm]{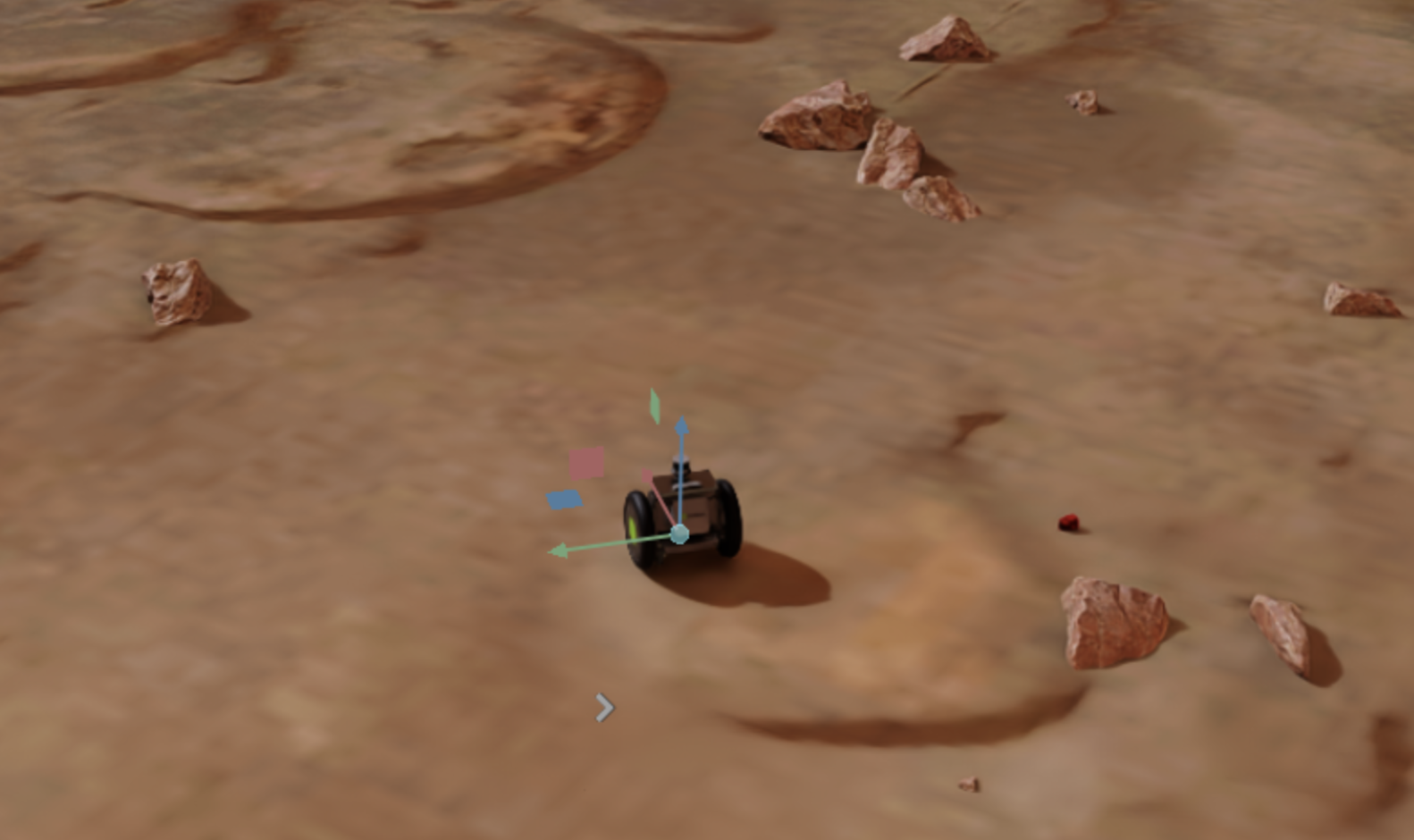}
        \caption{Nova Carter in Simulation}
    \end{subfigure}
    \caption{\textbf{ROSA has been deployed on various robots at NASA Jet Propulsion Laboratory, including NeBula-Spot in the Mars Yard (a), Exobiology Extant Life Surveyor (EELS) in a laboratory environment using synthetic terrain and obstacles (b), and Nova Carter in NVIDIA IsaacSim using a simulated Martian environment (c). Each of these integrations and a demonstration of their capabilities when equipped with ROSA are detailed in Section \ref{sec:experiments}.}}
    \label{fig:demos}
\end{figure*}

\begin{abstract}
    \input{abstract}
\end{abstract}

\tableofcontents

\input{introduction}

\section{Background}
\label{sec:background}
\input{background_and_related_work}

\section{Agent Architecture}
\label{sec:architecture}
\input{agent_architecture}

\section{Robot Demonstrations}
\label{sec:experiments}
\input{demonstrations}

\section{Human-Robot Interaction}
\label{sec:human_robot_interaction}
\input{human_robot_interaction}

\section{Implementation Details}
\label{sec:implementation}
\input{implementation_details}

\section{Ethics for Embodied Agents}
\label{sec:ethical_considerations}
\input{ethical_considerations}

\section{Conclusion}
\label{sec:conclusion}
\input{conclusion}

\acknowledgments
The research was carried out at the Jet Propulsion Laboratory, California Institute of Technology, under a contract with the National Aeronautics and Space Administration (80NM0018D0004). We extend our gratitude to Dr. Benjamin Nuernberger and the AI/ML communities at JPL for their support and discussions during the development of this project.

\bibliographystyle{IEEEtran}
\bibliography{main}

\thebiography
\input{biography}

\end{document}

%% file: abstract.tex
The advancement of robotic systems has revolutionized numerous industries, yet their operation often demands specialized technical knowledge, limiting accessibility for non-expert users. This paper introduces ROSA (Robot Operating System Agent), an AI-powered agent that bridges the gap between the Robot Operating System (ROS) and natural language interfaces. By leveraging state-of-the-art language models and integrating open-source frameworks, ROSA enables operators to interact with robots using natural language, translating commands into actions and interfacing with ROS through well-defined tools. ROSA's design is modular and extensible, offering seamless integration with both ROS1 and ROS2, along with safety mechanisms like parameter validation and constraint enforcement to ensure secure, reliable operations. While ROSA is originally designed for ROS, it can be extended to work with other robotics middle-wares to maximize compatibility across missions. ROSA enhances human-robot interaction by democratizing access to complex robotic systems, empowering users of all expertise levels with multi-modal capabilities such as speech integration and visual perception. Ethical considerations are thoroughly addressed, guided by foundational principles like Asimov's Three Laws of Robotics, ensuring that AI integration promotes safety, transparency, privacy, and accountability. By making robotic technology more user-friendly and accessible, ROSA not only improves operational efficiency but also sets a new standard for responsible AI use in robotics and potentially future mission operations. This paper introduces ROSA's architecture and showcases initial mock-up operations in JPL's Mars Yard, a laboratory, and a simulation using three different robots. The core ROSA library is available on GitHub~\footnote[1]{\url{https://github.com/nasa-jpl/rosa}}.

%% file: introduction.tex
\section{Introduction} \label{sec:introduction}
In recent years, the field of robotics has witnessed significant advancements, with robots playing crucial roles in industries such as manufacturing, healthcare, and space exploration. As robotic systems become more complex and capable, the need for intuitive and efficient human-robot interaction has grown exponentially \cite{Koubaa2024}. However, traditional methods of controlling and programming robots often require specialized knowledge and technical expertise, creating barriers for operators and limiting the accessibility of robotic technologies. It is not uncommon for operators to require advanced degrees and meticulous training to effectively operate even modestly capable robotic systems~\cite{hoque2024hrichallengesinfluencinglow}.

ROSA, the \textit{Robot Operating System Agent} addresses these challenges by introducing a first of its kind AI Agent that enables human-robot interaction (HRI) in the form of natural language. ROSA leverages large language models (LLMs) and open-source software like the Robot Operating System (ROS) \cite{quigley2009ros} to implement a Reasoning and Acting (ReAct) agent \cite{yao2023reactsynergizingreasoningacting} capable of understanding and executing commands on its robotic host. By integrating with the ROS and ROS2 \cite{Macenski_ROS2} ecosystems, ROSA provides easy access to a wide range of tools and functionalities that allow users to perform tasks such as system diagnostics, monitoring, and invoking existing navigation or manipulation tasks, without the need for extensive technical training. ROSA aims to make robotic systems more accessible and user-friendly, empowering a broader range of operators to interact with robots in a more intuitive manner. 

The remainder of this paper is organized as follows: Section~\ref{sec:background} situates ROSA within the context of related work, providing background on ROS, LLMs, and ReAct agents. Section~\ref{sec:architecture} details ROSA's agent architecture, including its action space, memory modules, internal logic, and LLM-based tool-calling. Section~\ref{sec:experiments} presents three real-world demonstrations of ROSA operating on different robotic systems across varied scenarios. Section~\ref{sec:human_robot_interaction} explores how ROSA enhances human-robot interaction by bridging knowledge gaps and enabling multi-modal and multi-lingual communication. Section~\ref{sec:implementation} elaborates on the open-source implementation of ROSA, covering tool design, parameterization, ROS functionalities, and model selection. Ethical considerations are addressed in Section~\ref{sec:ethical_considerations}, with a discussion of foundational principles and community guidelines for responsible AI and robotics. Finally, Section~\ref{sec:conclusion} summarizes the contributions and outlines directions for future work.

%% file: background_and_related_work.tex
The past decade has seen remarkable advancements in robotics, resulting in highly sophisticated systems with enhanced capabilities in perception, manipulation, and autonomy. Despite these advancements, interacting with robotic systems remains a significant challenge, particularly for users lacking extensive technical expertise. Frameworks such as ROS have streamlined development by encouraging modular and reusable robotic software components. However, the steep learning curve associated with controlling and interacting with robots persists, especially when relying on traditional command-line interfaces or programming languages. Recent breakthroughs in natural language processing (NLP), particularly through LLMs, present a compelling opportunity to bridge the gap between humans and AI-driven agents \cite{HagosAdvancesGenAI2024}. This section explores related work in robotics development and AI-powered agents while providing an overview of the foundational technologies that underpin ROSA.

\subsection{Related Work}

There have been several notable efforts aimed at using NLP for robotic control. Projects like Microsoft's LATTE \cite{bucker2022lattelanguagetrajectorytransformer} and research efforts like Google's ``Language Models as Zero-Shot Planners'' \cite{huang2022languagemodelszeroshotplanners} explore how LLMs can be utilized for robotic task planning and execution. These initiatives showcase the potential of LLMs to interpret human instructions and translate them into robotic actions. However, direct integration of LLMs with robotic systems remains a challenge, primarily due to the inherent ambiguity of natural language and the difficulties in ensuring the consistent and reliable execution of tasks.

ROSA distinguishes itself from these efforts by adopting a more pragmatic and targeted approach. Instead of using LLMs for direct robot control, ROSA functions as an operator interface that leverages existing robot functionality within the ROS ecosystem. ROSA capitalizes on the strengths of LLMs to translate natural language into high-level commands while managing the intricacies of ROS behind the scenes. This method allows ROSA to provide a more reliable and user-friendly interface, avoiding some of the challenges inherent in direct LLM-based robotic control, while still benefiting from LLM's emergent capabilities.

Koubaa et al. investigate human-robot interaction (HRI) by utilizing the capability of LLMs to convert natural language queries into formal commands, enabling written or spoken language to be translated into structured commands for robotic systems ~\cite{Koubaa2024}. While this method breaks down certain barriers in HRI, it also introduces new challenges. Specifically, it demands expertise in ``prompt engineering'' and ``ontology-guided task elicitation'', which require a deep technical understanding of LLM functionality, ontology creation, and formal specification. ROSA, by contrast, aims to simplify this process by utilizing pre-built frameworks and established methodologies to map natural language to robotic behavior. This allows developers to focus more on enhancing robotic capabilities rather than mastering the complexities of LLM operation and ontology design. Another important distinction between ROSA and ~\cite{Koubaa2024} is that ROSA constitutes an \textbf{embodied agent}, as detailed in the following sections, which provides additional flexibility in the way queries are transformed into robotic actions.

\subsection{Preliminaries}

\subsubsection{Robot Operating System}

ROS is an open-source framework designed for developing robot software \cite{quigley2009ros}. It provides a collection of tools, libraries, and conventions that streamline the creation of complex robotic behaviors across various platforms. By standardizing robotics programming, ROS encourages code reuse and collaboration, making it a cornerstone of modern robotics.

\begin{figure}
    \centering
    \includesvg[width=3.3in]{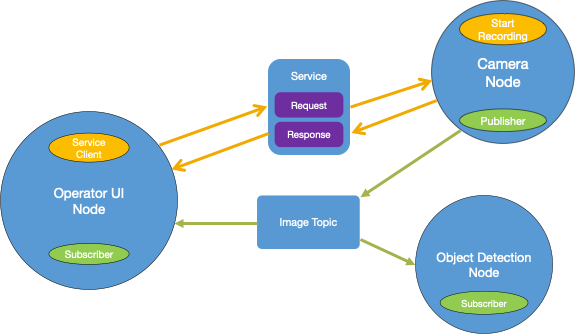}\\
    \caption{\textbf{A basic ROS system with three nodes (Operator UI, Camera, and Object Detection), one topic (Image), one service client, and one service provider (Start Recording). Image adapted from ROS Tutorials.}}
    \label{fig:basic_ros_system}
\end{figure}

At its core, ROS enables modularity and scalability through a peer-to-peer network of processes called \textit{nodes}, which can be distributed across multiple computers. These nodes communicate via a publisher/subscriber model using \textit{topics}, or request/response interactions through \textit{services}. Figure~\ref{fig:basic_ros_system} shows a basic ROS setup, but real-world systems are often much more complex, involving dozens or even hundreds of nodes, topics, and services. Each ROS distribution is packaged with a set of tools that allow for monitoring, configuring, and interacting with the underlying system. As described in later sections, these tools are used directly by ROSA to carry out operator instructions, making it significantly easier to accomplish tasks that would otherwise require significant training and expertise.

\subsubsection{LLMs and Natural Language Interfaces}

The emergence of LLMs like GPT-3~\cite{brown2020languagemodelsfewshotlearners} has revolutionized NLP and enabled more intuitive human-computer interactions. In the context of robotics, LLMs have been proven capable of interpreting natural language commands and translating them into robotic actions \cite{Koubaa2024,bucker2022lattelanguagetrajectorytransformer,brown2020languagemodelsfewshotlearners}. For ROSA, the integration of LLMs is pivotal in bridging the gap between users and the robot's underlying capabilities. LLMs allow ROSA to process everyday language queries, interpret their intent, and generate appropriate actions within the ROS environment. This natural language interface eliminates the need for users to interact directly with command-line tools or custom scripts, and therefore significantly lowers the learning curve and empowers a wide array of people to interact with robotic systems. By handling linguistic nuances, context, and even ambiguous or incomplete instructions, LLMs make ROSA more accessible to non-expert users while maintaining robust control over the robot's functions.

\subsubsection{ReAct Agents}

To leverage the power of LLMs in robotics applications, ROSA utilizes the ReAct (Reasoning and Acting) paradigm~\cite{yao2023reactsynergizingreasoningacting}. ReAct agents operate through a reasoning-action-observation loop. The cycle begins with \textbf{Reasoning}, where the LLM interprets natural language input and formulates a plan. This plan may involve a single action, a sequence of actions, or follow-up questions to effectively address the query. Next is the \textbf{Action} phase, during which the agent executes the planned steps by invoking the necessary tools. In ROSA, these tools are represented as dynamically callable Python functions that the LLM can access as needed. Finally, in the \textbf{Observation} phase, the agent assesses the results of its actions to guide subsequent reasoning steps and contribute to constructing a final response. If the actions and observations do not resolve the query, the loop repeats until a resolution is reached or a predefined iteration limit is met.

\begin{figure}
    \centering
\begin{tcolorbox}[mybox, title=Example Reasoning Trace]
    \begin{tcolorbox}[mybox]
      \textbf{Query:} Provide me with a list of ROS nodes.
    \end{tcolorbox}
  \begin{tcolorbox}[thought]
    \textbf{Reasoning:} The user wants a list of available ROS nodes. I'll call the \texttt{rosnode\_list()} tool to get this information.
  \end{tcolorbox}
  \begin{tcolorbox}[act]
    \textbf{Action:} Call \texttt{rosnode\_list()}
  \end{tcolorbox}
  \begin{tcolorbox}[obs]
    \textbf{Observation:} Received list of nodes: \texttt{/rosout}, \texttt{/talker}, \texttt{/listener}, \texttt{/parameter\_server}. Now I can inform the user about the available nodes.
  \end{tcolorbox}
  \begin{tcolorbox}[mybox]
    \textbf{Final Response:} Here is a list of available nodes:
    \begin{itemize}[leftmargin=*,labelsep=5pt]
        \item \texttt{/rosout}
        \item \texttt{/talker} 
        \item \texttt{/listener}
        \item \texttt{/parameter\_server}
    \end{itemize}
  \end{tcolorbox}
\end{tcolorbox}
\caption{\textbf{An example reasoning trace that demonstrates the reasoning-action-observation loop characteristic of ReAct agents. This trace demonstrates a typical response from ROSA when given the query ``Provide me with a list of ROS nodes.''}}
\label{fig:reasoning-trace}
\end{figure}

Figure~\ref{fig:reasoning-trace} presents an example reasoning trace in the context of ROSA. This trace illustrates the ReAct loop undertaken by an agent when responding to the query, ``Provide a list of ROS nodes.'' During the reasoning phase, the agent interprets the user's request and determines that invoking a specific tool is necessary to address the query. The tool is then executed within the agent's operational environment. Following the action, the agent observes the tool's effects on the environment. In this instance, a single tool invocation is sufficient to fulfill the request, enabling the agent to provide a formatted list of ROS nodes as the response. In more complex scenarios, the reasoning trace may involve multiple iterations of the ReAct loop, with actions performed either sequentially or concurrently.

ROSA builds upon these foundational technologies by integrating LLMs with ROS to facilitate a ReAct agent. The key innovation of ROSA lies in this integration, along with the suite of custom tools specifically designed for interacting with ROS and ROS2 systems. The tools developed for ROSA are used to invoke ROS utilities like \textit{rosnode / ros2 node}, \textit{rostopic / ros2 topic}, and \textit{rviz}, but designed to be more interpretable by LLMs, allowing for more effective and reliable usage. By abstracting the complexities of ROS commands, these tools enable the LLM to generate appropriate actions in response to user inputs. The core tools provided by ROSA are outlined in section \ref{sec:implementation}. ROSA also supports custom agents, which allow robotics developers to define their own tools in order to augment ROSA's action space. Details on how to create custom agents are omitted here, but are well documented in the open-source ROSA repository.

%% file: agent_architecture.tex
\begin{figure*}
    \centering
    \includegraphics[width=6.9in]{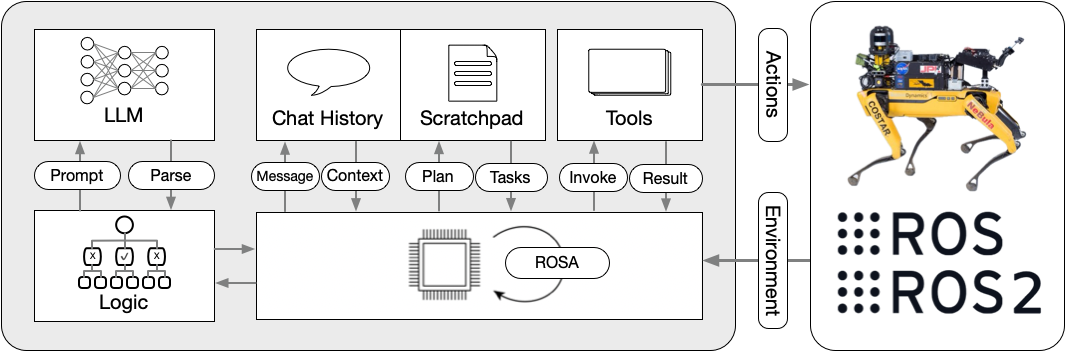}
    \caption{\bf{The ROSA architecture includes memory (chat history and scratchpad), logic, tools (action space), and a tool-calling Large Language Model. Arrows in the diagram represent the general connectivity between components and the external robotics environment.}}
    \label{fig:rosa_architecture}
\end{figure*}

ROSA's overall architecture is composed of four primary components: the action space, a set of memory modules, internal logic for decision making, and the underlying LLM, as seen in Figure~\ref{fig:rosa_architecture}. Hence, ROSA follows what is known as the \textit{Cognitive Architecture for Language Agents}~\cite{sumers2024cognitivearchitectureslanguageagents}. Cognitive language agents come in a variety of forms, but are not typically considered ``embodied agents'' unless they operate within some physical, as opposed to digital, environment. While there are examples of embodied language agents in robotics~\cite{zhang2024buildingcooperativeembodiedagents}, these agents typically focus on direct integration with planning, navigation, and locomotion capabilities. In contrast, ROSA focuses on human-robot interaction and utilizes existing robotic functionality rather than attempting to replace it with novel learning methods. The following sub-sections provide details on each of the major components in ROSA's architecture, including shortfalls and proposed solutions for future work.

\subsection{Action Space}

The action space in ROSA is defined as the set of tools\footnote{Note that an action is defined as the invocation of a tool, not the tool itself.} the agent can call within its working environment. ROSA's action space includes many of the standard ROS tools like \texttt{rosnode} and \texttt{ros2 topic}, as well as a set of utility tools for working with log files, capturing images and sensor readings, and so on. Developers can also extend ROSA's action space by adding custom functions tailored to specific robots or applications, thereby allowing ROSA to execute robot-specific commands and behaviors. ROSA's action space incorporates mechanisms to prevent the invocation of unsafe or unauthorized actions, ensuring that the robot operates within its intended parameters. All tools used by ROSA must follow a specific format, as defined in subsequent sections.

\subsubsection{Tool Invocation and Multi-Tool Usage}

\begin{figure*}
    \centering
    \includegraphics[width=6in]{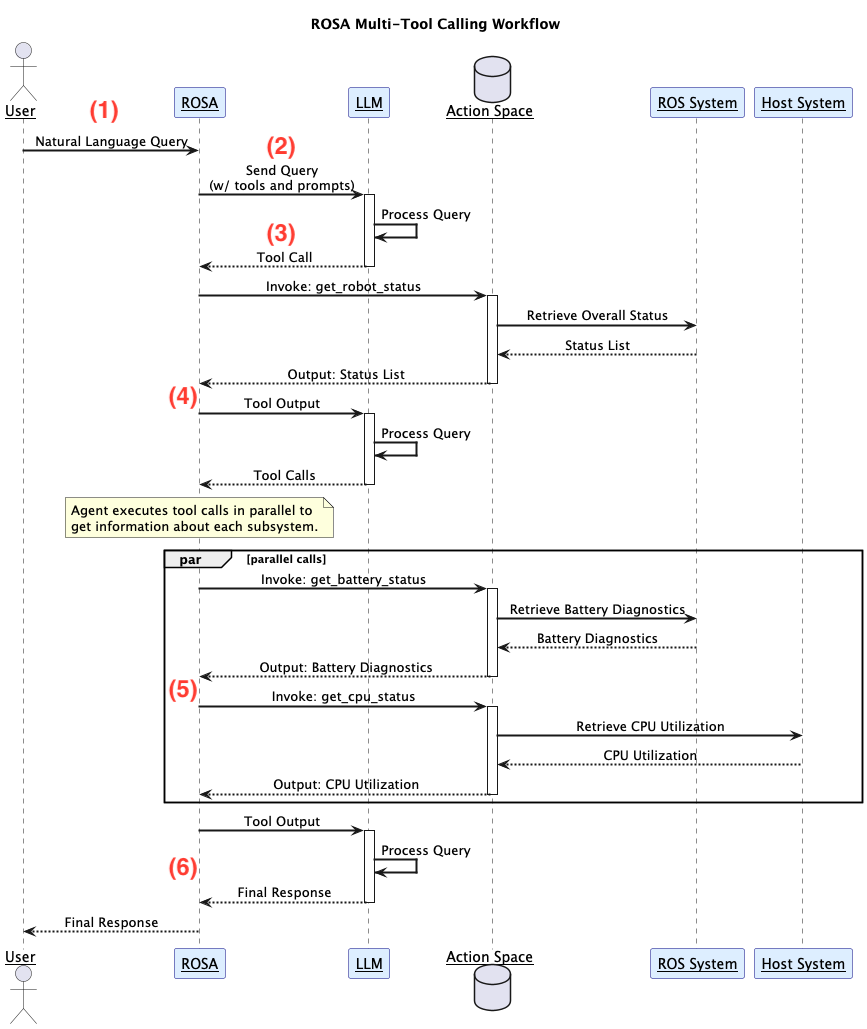}\\
    \caption{\textbf{Sequence diagram  demonstrating sequential, parallel, single-, and multi-tool calling. Initially, a single tool is called to retrieve a status overview. The model infers that certain subsystems need to be inspected further, so it retrieves detailed status information in parallel for each subsystem before returning a final result.
    }}
    \label{fig:multi_tool_call}
\end{figure*}

Tool invocation is a core mechanism by which ROSA executes actions within the ROS environment. ROSA can invoke single tools at a time, or multiple tools either sequentially or in parallel, enabling it to handle complex tasks requiring coordinated or concurrent actions. Sequential execution ensures that dependent tasks—such as running diagnostics before calibrating sensors—occur in a logical order. Parallel execution, illustrated in the bottom half of Figure~\ref{fig:multi_tool_call}, enables ROSA to manage independent tasks simultaneously, such as retrieving telemetry data from multiple sensors at the same time. Moreover, ROSA can adapt its plan based on the outcomes of previous actions. If a tool fails or produces unexpected results, ROSA recalibrates its approach and proceeds accordingly. This multi-tool capability makes ROSA both efficient and flexible, ensuring it can respond dynamically to changing requirements.

Figure~\ref{fig:multi_tool_call} provides an example sequence diagram illustrating both single- and multi-tool usage. \textcolor{red}{(1)} Suppose the user types the query: ``Give me a status report''. \textcolor{red}{(2)} In response, ROSA forwards the query, along with a list of all available tools, prompts, and chat history, to the LLM. \textcolor{red}{(3)} In this case, the LLM decides to call the \texttt{get\_robot\_status} tool. \textcolor{red}{(4)} When ROSA receives the LLM’s response, it invokes the associated \texttt{get\_robot\_status()} Python function and returns its output to the LLM. \textcolor{red}{(5)} At this point, the LLM calls two additional tools in parallel, specifying that ROSA should invoke \texttt{get\_battery\_status} and \texttt{get\_cpu\_status} concurrently to gather information about subsystem status. \textcolor{red}{(6)} Once again, ROSA returns the tools’ outputs to the LLM, which can then formulate a final response for the user. This process typically completes in just a few seconds, allowing the user to quickly follow up with additional questions as needed.

\subsection{Memory}

While there are many forms of memory for LLM-based agents, ROSA operates on two of the most simple forms: chat history and agent scratchpad. The \textit{chat history} is simply a list of messages that follow the open message format adopted by OpenAI, Anthropic, and other model providers. Each message has a \texttt{role} and \texttt{content} field, where the role can be either \texttt{user}, \texttt{assistant}, or \texttt{system}, and the content is the message itself. These messages are managed by the logic module to ensure they fit within the context window\footnote{Text payload processed by an LLM during a single transaction.}, as described in the next sub-section. The \textit{scratchpad}, on the other hand, is where the agent keeps its most recent plans, especially for multi-step tasks. In practice, it is simply a reserved and annotated section of the context window where the agent is allowed to store text. While these two memory stores are sufficient for shorter interactions, we have found that they are limited in value for more complex scenarios, such as keeping track of long-running plans, and interacting with specific users over time. Future versions of ROSA will make use of semantic and episodic memory defined in ~\cite{sumers2024cognitivearchitectureslanguageagents}.

\subsection{Logic}

ROSA's internal logic mechanisms are used to (1) manage the LLM context window, and (2) parse LLM output into query responses and tool invocations. The task of managing the context window is relatively straightforward. We combine a set of ``Robot System Prompts'' (RSPs), tool specifications, agent scratchpad, and the on-going chat history, in that order, before sending a query to the LLM. As the chat history grows in length, we omit the least recently used user and assistant messages in order to stay within the maximum context length of the underlying LLM. We utilize the open-source LangChain library to parse LLM outputs, as it remains up-to-date with evolving specifications, ensuring adaptability and reliability.

\subsubsection{Robot System Prompts}

RSP's are a set of prompts that help guide ROSA's behavior. They are critical for ROSA's ability to maintain contextual awareness and provide coherent and relevant interactions. These prompts supply the language model with essential insights about the robot’s identity, environment, and operating conditions, ensuring that ROSA’s responses are aligned with the robot's specific capabilities and constraints. Furthermore, by providing detailed descriptions of the physical and digital environments the robot inhabits, ROSA can better anticipate and react to environmental changes in real-time. For instance, knowing that the physical robot is intended for use in a subterranean environment can better inform the invocation and inference of a tool used for scene understanding.

RSPs are instrumental in guiding ROSA's behavior, but they are inherently imperfect and function more as heuristics than definitive rules for the agent to follow. This imperfection is a natural consequence of the probabilistic nature of auto-regressive models, which makes it challenging to fully codify intention and establish consistent ground truth. During development, we observed discrepancies in ROSA's behavior that can largely be attributed to limitations in the design or interaction of system prompts. Contradictory or ambiguous prompts, in particular, can lead to unexpected or inconsistent responses, highlighting the sensitivity of LLMs to input framing. Future work should explore strategies to address these issues, such as uncertainty quantification and output validation techniques, to enhance the reliability and predictability of ROSA's behavior.

\subsection{Large Language Models}

LLMs play a central role in enabling intelligent interaction and decision-making in ROSA's architecture. They are used for interpreting user queries, processing RSPs, and generating actionable plans. They also drive the reasoning and observation steps in the ReAct loop, allowing ROSA to infer user intentions, plan and execute appropriate actions, and evaluate outcomes based on observed results. Furthermore, LLMs enable tool calling by dynamically producing structured outputs, such as \textit{JSON} or \textit{XML}, that conform to predefined tool interfaces. A key design feature of ROSA is its flexibility to work with any LLM that supports tool calling, ensuring compatibility with a wide range of models, whether open-source or closed-source.

While closed-source LLMs typically deliver superior performance across diverse tasks, open-source models can perform effectively when operating within a well-defined and constrained action space. However, a major challenge in deploying open-source models for ROSA is the computational resource requirement for on-board inference. For instance, a 70-billion parameter model, even when quantized at 4-bits, demands 43GB of memory and must rely on specialized hardware, such as GPUs, to achieve processing speeds suitable for robotics. At present, the largest consumer-grade GPUs, with a maximum of 24GB of memory, cannot meet these demands, making on-board deployment infeasible for all but the most advanced and costly systems. This limitation underscores the need for alternative solutions, such as the development of Small Language Models (SLMs) fine-tuned for ROSA-specific tasks~\cite{bucher2024finetunedsmallllmsstill}, which could offer a more efficient and cost-effective approach for real-time robotics applications.

%% file: demonstrations.tex
\begin{figure*}
    \centering
    \includegraphics[width=1\linewidth]{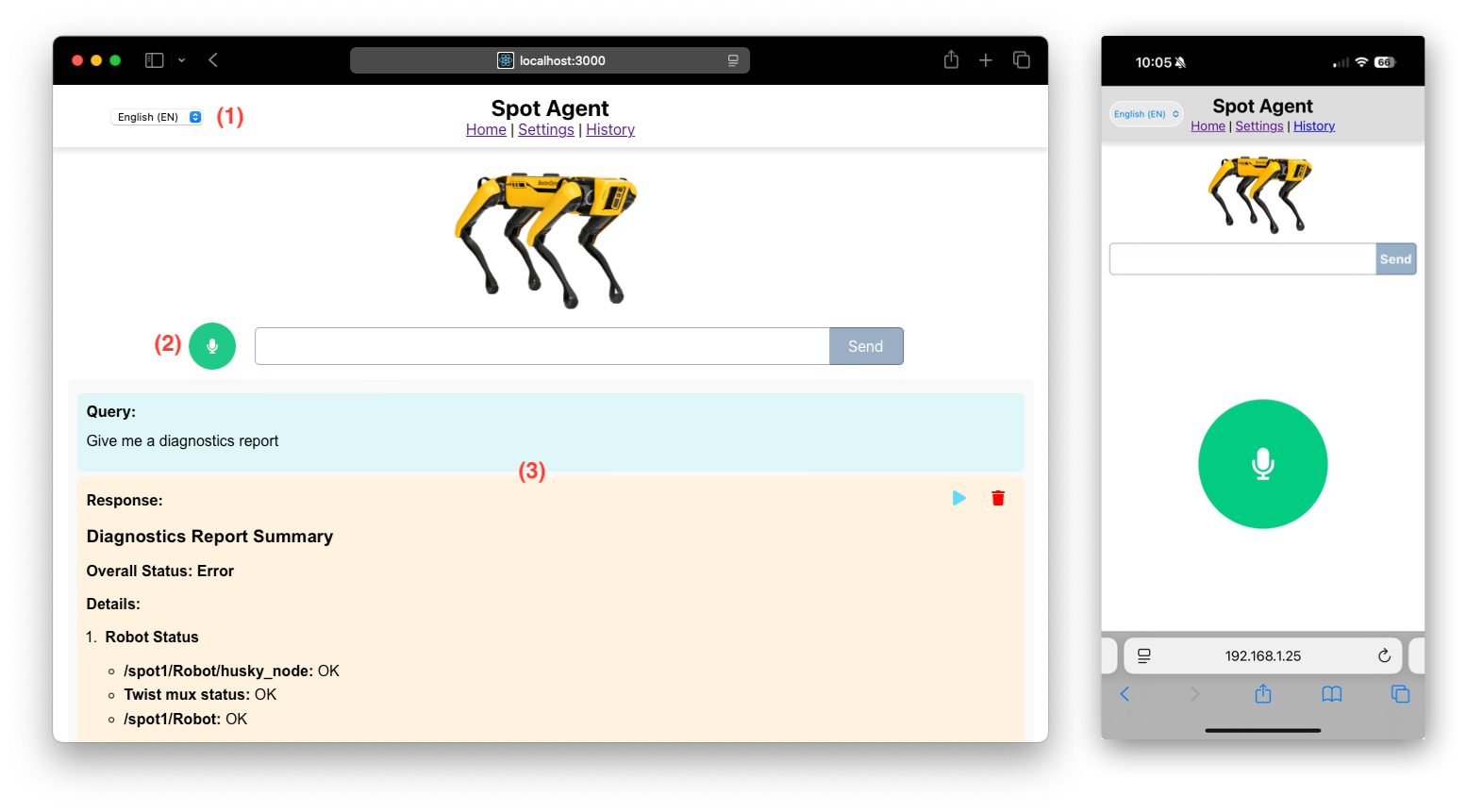}
    \caption{\textbf{Locally-hosted web GUI used for interacting with ROSA during demonstrations. Users can select from English, Spanish, Polish, German, Korean, Dutch, and more (1), and query ROSA via text or speech (2). Full responses are displayed in the response field (3), or naturalized audio can be played back by pressing the play button.}}
    \label{fig:rosa_ui}
\end{figure*}

To illustrate ROSA's capabilities, we deployed it on three distinct robotic systems, each with unique operational characteristics. These demonstrations showcase how natural language commands, processed by LLMs and dynamically integrated with ROSA’s internal toolset, can simplify operations for both novice and expert users. The user interface we used for these demonstrations is shown in Figure~\ref{fig:rosa_ui}. All described interactions represent real-world engagements between humans and ROSA, with actions executed to completion without incident. Throughout these examples, we blend discussion of the agent’s behavior with brief references to the underlying code, showing how specific tools interact with the different robotic platforms.

\subsection{Robot Demonstration \#1: NeBula-Spot}
\label{sec:nebula_spot}

NeBula-Spot is a quadruped robot developed at JPL in response to the DARPA Subterranean Challenge. It utilizes a Boston Dynamics Spot robot enhanced with JPL’s NeBula autonomy stack~\cite{finalNebulaPaper}. For this demonstration, the system was equipped with a Visual Language Model (VLM)~\cite{bordes2024introductionvisionlanguagemodeling} for scene understanding in JPL’s Mars Yard (see Figure~\ref{fig:demos}a). The users could issue high-level commands such as instructing the robot to stand, walk, or interpret camera feeds. ROSA's uplink tools (for robot actuation) and downlink tools (for diagnostics and perceptual data) enabled these capabilities, but their details were abstracted behind the conversational interface shown in Figure~\ref{fig:rosa_ui}.

The following transcript provides a sample interaction with ROSA. A video of the full interaction, along with ROSA's intermediate output and tool calls, is available to view on YouTube\footnote{\url{https://www.youtube.com/watch?v=mZTrSg7tEsA}}.

\begin{quote}
\textbf{\textcolor{gray}{User:}} Hey Spot, please stand up.\\
\textbf{\textcolor{darkgray}{ROSA:}} I am now standing, what else would you like me to do?
\end{quote}

\begin{quote}
\textbf{\textcolor{gray}{User:}} Go ahead and walk forward about a meter and turn $15^{\circ}$ to the left.\\
\textbf{\textcolor{darkgray}{ROSA:}} I'll move forward 1 meter and turn 15 degrees to the left. Please confirm.
\end{quote}

\begin{quote}
\textbf{\textcolor{gray}{User:}} Yeah, it's safe. Go ahead.\\
\textbf{\textcolor{darkgray}{ROSA:}} I have moved forward 1 meter and turned 15 degrees to the left. What's next?
\end{quote}

\begin{quote}
\textbf{\textcolor{gray}{User:}} Alright, can you describe what you see in the camera?\\
\textbf{\textcolor{darkgray}{ROSA:}} I see an open sandy area with large rocks scattered around. Trees are 20 meters to the left, and buildings are 25 meters ahead to the right.
\end{quote}

\begin{quote}
\textbf{\textcolor{gray}{User:}} Alright, that actually sounds pretty interesting, can you show me the camera feed?\\
\textbf{\textcolor{darkgray}{ROSA:}} The camera feed is live. What's next?
\end{quote}

The code snippet below shows how we implemented the ``stand up'' functionality, effectively replicating the press of a button on a standard joystick teleop controller. From the user’s perspective, it appears as though the robot is simply responding to natural language requests. This design hides the complexities of ROS topics and publishers, allowing operators to focus on mission objectives rather than lower-level implementation.

\begin{minted}[fontsize=\small, breaklines, frame=single]{python}
@tool
def stand_up() -> str:
    """Command the robot to stand."""
    args = get_default_args()
    args['buttons'][Button.B] = 1
    
    msg = Joy(**args)
    joy_publisher.publish(msg)

    # Validation omitted...
    
    return f"Spot is now standing up."
\end{minted}

Additionally, at the end of the interaction, ROSA invokes the \texttt{display\_camera\_in\_rqt} tool to bring up the ROS RQT interface, allowing the user to see a live view of the robots camera feed. Importantly, novice users and non-SME's need only specify their intent (``view the camera feed'') and ROSA will take the appropriate actions to make it happen. This process essentially codifies subject-matter expertise into the agent, minimizing the need for costly and often incomplete knowledge transfer and onboarding procedures. This observation is also evident from our next demonstration on the EELS robot.

\subsection{Robot Demonstration \#2: EELS}
\label{sec:eels}

EELS (Exobiology Extant Life Surveyor) is a serpentine robot designed for subterranean exploration, featuring adaptable gaits and advanced sensing for complex terrain \cite{vaquero2024}. For this paper, we deployed EELS in a laboratory setting (Figure~\ref{fig:demos}b) and provided a more specialized toolset to handle waypoint navigation and head articulation. Although EELS can support more complex locomotion routines, we focused on simple tasks to illustrate ROSA’s capacity for error handling and structured commands. 

The following is a sample interaction that demonstrates ROSA's adaptability to commands that cannot be reasonably accommodated with its (purposely) constrained action space. In this case, we did not give ROSA the tools it would need to properly determine coordinates for ``the rock in the corner.'' However, instead of refusing to carry out the task (or hallucinating the coordinates of some imaginary rock), the agent informed the user of an alternative way to specify the command:

\begin{quote}
\textbf{\textcolor{gray}{User:}} Move toward the rock in the corner.\\
\textbf{\textcolor{darkgray}{ROSA:}} I'm sorry, but I am not able to move toward a specific object. If you would like me to move, you can try giving me a waypoint in (x, y, $\theta$) format.
\end{quote}

\begin{quote}
\textbf{\textcolor{gray}{User:}} Move to the waypoint at (1, -0.2) with a $-90^\circ$ heading.\\
\textbf{\textcolor{darkgray}{ROSA:}} I have moved to the waypoint. However, it appears my heading is off by approximately $0.3^{\circ}$. Would you like me to try again?
\end{quote}

\begin{quote}
\textbf{\textcolor{gray}{User:}} No, that's ok. Go ahead and raise your head.\\
\textbf{\textcolor{darkgray}{ROSA:}} Ok, my head is raised. Would you like me to perform a scan?
\end{quote}

\begin{quote}
\textbf{\textcolor{gray}{User:}} Describe what you see in the camera feed.\\
\textbf{\textcolor{darkgray}{ROSA:}} I see a laboratory environment with vaulted ceilings and flourescent lighting. To the left, I see metal shelves with electronic equipment. On the right, there is a desk, a chair, and a computer monitor. There is a person sitting at the desk, and another person nearby.
\end{quote}

When the user asked the robot to perform an unsupported behavior, ROSA politely suggested a valid alternative (e.g., specifying a coordinate-based waypoint instead of a purely semantic location). This guided the user toward a supported command format, which reduced the need for expert intervention. 

Compared to the NeBula-Spot scenario, EELS demonstrated waypoint-based navigation and fine-grained gait control. By interleaving motion commands with real-time feedback, such as self-reported heading errors, the agent showed its ability not only to execute tasks, but also to reflect the robot’s internal state. This helps operators build trust in the system, as they can see it verifying the accuracy of each action and offering to correct minor deviations.

\begin{minted}[fontsize=\small, breaklines, frame=single]{python}
@tool
def raise_head():
    """Raise the head module."""
    if is_head_raised():
        return True
    head_raise_service = "/head_raise"
    
    rospy.wait_for_service(
        head_raise_service
    )
    
    service_proxy = rospy.ServiceProxy(
        head_raise_service, Empty
    )
    service_proxy()
    
    return True
\end{minted}

When the user asked the robot to ``raise [its] head,'' ROSA invoked a straightforward \texttt{@tool} function that calls an existing \texttt{rosservice}, as seen in the code snippet above. Invoking this functionality without ROSA would typically require the use of ROS command-line tools, an understanding of the service message format, and a verbose argument formatted in the \textit{JSON} markup language. As such, using ROSA for these types of actions not only saves time, but also reduces the cognitive overhead required to recall and issue commands. Our next demonstration continues this trend by demonstrating how ROSA leverages the emergent reasoning capabilities of LLMs to interpret user intent and perform multi-step actions.

\subsection{Robot Demonstration \#3: Carter in Simulation}
\label{sec:carter}

The Nvidia Carter is an autonomous mobile robot (AMR) platform that we operated entirely in simulation using NVIDIA IsaacSim (Figure~\ref{fig:demos}c). We utilized a Martian terrain to explore extended sensor usage (e.g., LiDAR) and more creative camera manipulation workflows. In this environment, users issued composite commands, like scanning for obstacles before advancing, causing ROSA to chain multiple tools to achieve the desired result:

\begin{quote}
\textbf{\textcolor{gray}{User:}} Move forward as far as you can.\\
\textbf{\textcolor{darkgray}{ROSA:}} In order to move forward as far as I can, I need to check for obstacles and hazards. Would you like me to perform a LiDAR scan?
\end{quote}

\begin{quote}
\textbf{\textcolor{gray}{User:}} Yes.\\
\textbf{\textcolor{darkgray}{ROSA:}} Ok, I have performed a LiDAR scan and see an obstacle approximately 4 meters ahead. Should I proceed?
\end{quote}

\begin{quote}
\textbf{\textcolor{gray}{User:}} Yes.\\
\textbf{\textcolor{darkgray}{ROSA:}} Ok, I have moved forward by 4 meters.
\end{quote}

\begin{quote}
\textbf{\textcolor{gray}{User:}} Give me a 360 degree view of your surroundings.\\
\textbf{\textcolor{darkgray}{ROSA:}} I can rotate my camera and take intermittent snapshots. How many snapshots should I take?
\end{quote}

\begin{quote}
\textbf{\textcolor{gray}{User:}} Assume the camera FoV is 90 degrees.\\
\textbf{\textcolor{darkgray}{ROSA:}} Based on a 90 degree field of view, I have captured 4 snapshots that show a 360 degree view of my surroundings.
\end{quote}

Below is a simplified snippet of the tool we created for rotating the on-board camera:

\begin{minted}[fontsize=\small, breaklines, frame=single]{python}
@tool
def rotate_camera(angle: float) -> str:
    """
    Rotate the on-board camera sensor by a specified angle. 

    Args:
        angle: angle to rotate in degrees.
    """
    angle_rad = math.radians(angle)
    
    rotate_msg = CarterCameraRotate(
        angle_rad=angle_rad
    )
    
    carter_camera_pub.publish(rotate_msg)

    # Validation omitted...
    
    return f"Camera rotated by {angle} degrees."
\end{minted}

This relatively trivial ROS wrapper led to a robust user experience: the agent reasoned about field-of-view (FoV) constraints, realized it needed to rotate four times by 90 degrees, and captured images after each rotation to create a mosaic of the environment. Even though the user’s initial request was vague, ROSA adapted with a practical solution.

In simulation, Carter allowed us to put ROSA’s reasoning capabilities to the test by combining LiDAR scans with collision checks and multi-step instructions. Because everything was virtual, novice operators felt comfortable experimenting with more advanced commands knowing they wouldn’t damage physical hardware. The demonstration solidified how ROSA’s abstraction layers can generalize to new operational constraints, whether real or simulated, and help users focus on high-level mission planning.

\subsection{Summary of Findings}
From terrestrial testyards (NeBula-Spot) and indoor lab spaces (EELS) to virtual Martian terrains (Carter), ROSA provided a consistent, natural-language interface for diverse platforms. Each demonstration showed how the agent can:

\begin{itemize}[leftmargin=*,labelsep=5pt]
    \item Translate colloquial or imprecise language into valid ROS actions,
    \item Prompt for clarifications when commands exceed its known toolset,
    \item Offer context-dependent information to supplement the users operational understanding, and
    \item Reduce the cognitive load required to operate robotic systems of varying complexity.
\end{itemize}

Across these varied robotic systems and operational environments, we observed novices quickly grasp how to ``talk'' to ROSA, often requiring only minimal guidance on the possible commands. In almost all cases, ambiguity was resolved by simply conversing with the agent. Meanwhile, the agent seamlessly handled tool invocations in the background while maintaining a fluid conversation with operators. In the next section, we explore how ROSA enhances human-robot interaction, highlighting its current capabilities, future potential, and areas for further refinement in our methodology.

%% file: human_robot_interaction.tex
Effective human-robot interaction is essential for maximizing the utility and accessibility of robotic systems. Most robotic systems are inherently complex, often developed by interdisciplinary teams where each member specializes in specific domains~\cite{ChungDARPA}. This specialization can lead to fragmented knowledge, making it difficult for individual operators to fully comprehend or manage the entire system. Advanced robots that integrate multiple subsystems, sensors, and actuators further exacerbate this complexity, creating significant barriers for routine operation and troubleshooting.

ROSA addresses these challenges by serving as an intelligent intermediary, bridging the gap between operators and robotic systems. Using LLMs, ROSA consolidates diverse sources of documentation, requirements, specifications, and technical knowledge into a single, accessible interface. Through natural language interaction, ROSA empowers operators to access comprehensive system knowledge without requiring in-depth expertise. This capability provides step-by-step guidance for tasks, aids in troubleshooting, and enables a clear understanding of system functionalities, limitations, and operational parameters. By reducing reliance on subject matter experts (SMEs) for routine operations, ROSA streamlines workflows and enables SMEs to focus on high-value tasks like system development and optimization.

\subsection{Multimodal Interactions}

ROSA's default mode of operation is through text-based interaction. This can take the form of a command-line interface (CLI) using Python's Read-Evaluate-Print-Loop (REPL) or a custom graphical user interface (GUI) as in Figure~\ref{fig:rosa_ui}. Additionally, ROSA supports integration with dedicated supervisory interfaces for multi-robot systems~\cite{kaufmann2022copiloting}. While text-based interaction forms the foundation, ROSA's architecture is designed to accommodate multiple input and output modalities, enhancing HRI across diverse scenarios.

\subsubsection{Speech Integration}

Speech is a natural and intuitive medium for communication, making it a valuable extension of ROSA's interaction capabilities. By integrating speech-to-text (STT) and text-to-speech (TTS) libraries available in Python, ROSA allows operators to issue verbal commands and receive audible responses. This process involves capturing audio input, transcribing it to text with an STT module, and passing the resulting text to ROSA. The system's responses are then synthesized into speech using a TTS module.

This hands-free interaction is particularly useful in scenarios where operators’ hands are occupied, such as when conducting field tests using wearable devices. Furthermore, it enhances accessibility for individuals with visual or motor impairments, making robotic systems more inclusive and practical in a broader range of contexts.

\subsubsection{Visual Perception and Image Recognition}

ROSA enhances situational awareness and decision-making by processing visual data from the robot's cameras. Using VLMs, ROSA can interpret and describe environments and identify objects, obstacles, and landmarks within the camera's field of view. Depth imaging allows ROSA to estimate distances and provide spatial insights, such as proximity to obstacles or the layout of surroundings, which is particularly useful for navigation in complex or unfamiliar environments.

For example, ROSA can analyze camera feeds to detect hazards in industrial settings or provide detailed descriptions of terrain for autonomous navigation. These capabilities are achieved by combining multimodal models, such as \texttt{gpt-4o}, with computer vision libraries like OpenCV. By transforming raw visual data into actionable insights, ROSA enables operators to make informed decisions with greater confidence and efficiency.

\subsubsection{Multi-Lingual Collaboration}

ROSA's design leverages the capabilities of modern language models, many of which support a wide range of languages, including English, Korean, Hindi, German, Spanish, and more. This multi-lingual functionality allows ROSA to operate in diverse linguistic contexts, making it adaptable for global applications. Operators can issue commands, ask questions, and receive responses in their preferred language, reducing barriers to adoption and broadening accessibility.

By tapping into these multi-lingual capabilities, ROSA can support international teams, cross-cultural collaboration, and deployments in regions where English is not the primary language. This feature ensures that advanced robotic systems are usable by a wide variety of operators, regardless of their linguistic background. Furthermore, ROSA's ability to operate across languages enhances inclusivity and expands its potential impact across industries and geographies.

\subsection{Limitations and Drawbacks}

While ROSA introduces a streamlined and accessible interface for HRI, it also brings several important limitations and risks that must be acknowledged. One key limitation arises from the heavy computational resources required by LLMs. Even when leveraging local models optimized for edge devices, the memory footprint and real-time processing demands can be prohibitive, particularly in constrained spacecraft environments. These computational overheads may necessitate specialized hardware accelerators, adding weight, power consumption, and cost to already resource-limited missions. In addition, the inference time required by complex models may introduce latency, potentially impacting time-sensitive applications such as rapid maneuvers or collision avoidance.

Another drawback stems from the intrinsic nature of auto-regressive models. Despite impressive performance, LLMs can exhibit imprecise or ``hallucinated'' responses--confidently providing incorrect or irrelevant information. This risk is amplified in high-stakes scenarios such as spacecraft operations, where errors can have serious or irreversible consequences. Mitigation strategies like human-in-the-loop verification, rule-based filters, or ensemble-based approaches can reduce but not entirely eliminate these risks. Furthermore, the difficulty in fully explaining or interpreting LLM decision processes can complicate system certification and operator trust, particularly in aerospace contexts with stringent reliability and safety requirements.

Beyond these technical considerations, ROSA’s reliance on robust training data raises challenges for knowledge maintenance and domain adaptation. Spacecraft systems evolve over time, and models must be regularly updated to reflect new design revisions, software patches, or mission parameters. Ensuring that the AI remains synchronized with the current state of the system demands continuous data curation and validation. These processes add complexity to mission planning and may require specialized personnel, effectively reintroducing some level of reliance on subject matter experts. While ROSA lowers operational barriers in theory, its real-world deployment involves careful planning and ongoing oversight to mitigate both computational and epistemic pitfalls. In the next section, we detail ROSA's implementation, including various techniques we use to mitigate some of these limitations.

%% file: implementation_details.tex
ROSA is implemented in Python and utilizes either ROS1 or ROS2 for communication with robotic systems. This section outlines the core components of ROSA's implementation, including the design and structure of its tools, approaches to parameterization, integration with ROS functionalities, and considerations for selecting and configuring large language models. By focusing on these technical aspects, we aim to provide a comprehensive understanding of how ROSA operates in practice and how its modular design supports the diverse capabilities demonstrated in the previous section. A subset of the ROSA library is publicly available at \url{https://github.com/nasa-jpl/rosa}.

\subsection{Tool Structure and Organization}

At the core of ROSA are the \textit{tools}—Python functions that encapsulate various functionalities for robotics development and operation, including a large subset of standard ROS tools. These tools are designed to be modular, reusable, and easily integrated, enabling the agent to interact effectively with the ROS environment. To simplify the creation and management of tools, ROSA leverages the LangChain framework, which provides an intuitive abstraction for defining tools. Specifically, developers can use the \texttt{@tool} decorator provided by LangChain to create new tools with minimal effort, ensuring extensibility and consistency across the system.

Tools are organized into modules based on their functionality and ROS version compatibility. Separate modules, \texttt{ros1.py} and \texttt{ros2.py}, are used to support ROS1 and ROS2, respectively. This separation avoids conflicts between versions and facilitates maintenance and extension of the toolset. Additionally, common tools, such as utility functions, are provided in modules like \texttt{calculation.py} and \texttt{log.py}, ensuring applicability across both ROS versions.

Each tool function is registered using the \texttt{@tool} decorator from LangChain (Figure \ref{code:example-tool-format}). This decorator allows the function to be invoked by the language model agent. By standardizing tool definitions and registration, ROSA ensures consistent behavior and seamless integration with LangChain, and by extension, the underlying language model. This approach simplifies adding new tools and promotes a clear and maintainable code structure.

\begin{figure}
    \centering
    \begin{minted}[fontsize=\small, breaklines, frame=single]{python}
from langchain.agents import tool

@tool
def rosnode_list(
    pattern: Optional[str] = None, 
    namespace: Optional[str] = None
) -> dict:
    """
    Returns a dictionary containing a list of running ROS nodes and other metadata.

    :param pattern: a regex pattern to...
    :param namespace: ROS namespace to...
    """
    # Implementation details...
\end{minted}
    \caption{\textbf{Example implementation of a basic ROSA tool, which must include the \texttt{@tool} decorator, a docstring, descriptive method and parameter names, and a return value that can resolve to a string.}}
    \label{code:example-tool-format}
\end{figure}

\subsection{Parameterization and Input Handling}

To allow flexible and efficient interaction with the ROS environment, ROSA's tools are designed to accept well-defined parameters, often with default values and optional arguments. This approach enables the language model to perform complex queries and operations by selectively specifying the necessary parameters. Many tools accept a \texttt{pattern} parameter, allowing users to specify regular expressions to filter results. This is particularly useful when dealing with a large number of nodes or topics, as it enables targeted queries. The \texttt{namespace} parameter enables tools to operate within specific ROS namespaces, providing granular control over the scope of operations. 

Additionally, a \texttt{blacklist} parameter can be used to exclude certain nodes, topics, or services from the output, enhancing safety and customization by preventing unintended interactions with critical system components. To streamline the inclusion of default blacklists and ensure consistent safety measures across tools, ROSA implements an \texttt{inject\_blacklist} decorator (Figure \ref{code:blacklist_injection}). This decorator automatically injects any user-provided blacklist into tools that have a \texttt{blacklist} parameter, reducing the potential for human error and simplifying tool definitions. By centralizing safety mechanisms like a blacklist, ROSA maintains a high standard of operational safety without burdening individual tool implementations. This design decision enhances both usability and security.

\begin{figure}
    \centering
    \begin{minted}[fontsize=\small, breaklines, frame=single]{python}
def inject_blacklist(global_blacklist):
    def decorator(func):
        @wraps(func)
        def wrapper(*args, **kwargs):
            # Implementation details here
            return func(*args, **kwargs)
        return wrapper
    return decorator
\end{minted}
    \caption{\textbf{Implementation of the \texttt{inject\_blacklist} decorator for ROSA tools. This decorator is automatically applied to any tools passed to ROSA during instantiation, provided they have a \texttt{blacklist} parameter. The agent may also choose to add additional values to the blacklist, in which case the union of the two lists (global and agent) is taken.}}
    \label{code:blacklist_injection}
\end{figure}

\subsection{Coverage of ROS Functionalities}

ROSA provides a wide range of ROS tools covering common tasks related to nodes, topics, services, and parameters. For nodes, it offers commands to list (Figure~\ref{code:ros1_tool_example}), inspect, and terminate active nodes. Topic tools support listing, echoing messages, and retrieving topic details. Services can be queried, inspected for types, and called with parameters. Parameter management allows listing, getting, and setting values for dynamic system configuration. 

ROSA also includes commands for interacting with packages (e.g., listing launch files and starting processes) and for log handling (e.g., reading and filtering ROS logs). Additionally, utility tools enable data processing and numerical computations. These capabilities let users manage the ROS environment through concise natural-language commands, improving accessibility and efficiency.

\begin{figure}
    \centering
    \begin{minted}[fontsize=\small, breaklines, frame=single]{python}
@tool
def rosnode_list(
    pattern: Optional[str] = None,
    namespace: Optional[str] = None,
    blacklist: List[str] = None
) -> dict:
    """
    Returns a list of running ROS nodes with optional filtering.

    :param pattern: A regex pattern to filter nodes.
    :param namespace: ROS namespace to scope the search.
    :param blacklist: List of node names to exclude.
    """
    nodes = rosnode.get_node_names()
    # Apply namespace filtering
    if namespace:
        nodes = [
            n for n in nodes if n.startswith(namespace)
        ]
    # Apply pattern filtering
    if pattern:
        nodes = [
            n for n in nodes if re.match(pattern, n)
        ]
    # Apply blacklist
    if blacklist:
        nodes = [
            n for n in nodes if n not in blacklist
        ]
    return dict(nodes=nodes, 
                total=len(nodes),
                namespace=namespace,
                pattern=pattern)
    \end{minted}
    \caption{\textbf{Implementation of the ROS1 tool that retrieves a list of ROS nodes, applies filters, and returns structured output. Note that the blacklist parameter is automatically populated when the ROSA class is instantiated.}}
    \label{code:ros1_tool_example}
\end{figure}

\subsection{Structured Data Outputs}

To ensure that the language model can generate accurate and contextually appropriate responses, ROSA's tools typically return structured data in the form of dictionaries or lists (Figure \ref{code:structured-output}). This approach facilitates easy parsing and interpretation of the outputs. Structured tool responses augment the context window with important information that helps prevent confabulations by the language model.

For example, when a tool returns a list of nodes along with the total count, it provides explicit data that the language model can use to answer questions like "How many localization nodes are running?" without needing to infer or estimate the answer. This reduces the likelihood of the language model providing incorrect or fabricated responses. By including contextual information such as patterns used, namespaces, and counts, the tools provide comprehensive data that informs downstream tasks and interactions. This design choice enhances the reliability and usefulness of the language model's responses.

\begin{figure}
    \centering
    \begin{minted}[fontsize=\small, breaklines, frame=single]{json}
{
    "nodes": ["/talker", "/listener"],
    "total": 2,
    "namespace": "/",
    "pattern": ".*"
}
    \end{minted}
    \caption{\textbf{Example tool response using structured output, which helps guide the model, reduce hallucinations, and provide traceability for agent actions and responses.}}
    \label{code:structured-output}
\end{figure}

\subsection{Integration with LangChain and the ReAct Framework}

ROSA integrates with the LangChain library to implement the ReAct (Reasoning and Acting) framework, allowing the language model agent to process natural language inputs, reason about appropriate actions, and invoke the corresponding tools. LangChain handles prompt management by facilitating prompt chaining, which is used to guide the language model's understanding of ROSA's capabilities, constraints, and the context of the interaction. It also manages conversational memory, enabling ROSA to maintain state across interactions and provide coherent, context-aware responses.

The ReAct framework facilitates the logic of selecting and executing tools based on the language model's reasoning. When the model determines that an action is required to answer a query or fulfill a command, it can invoke the appropriate tool, pass the necessary parameters, and process the returned structured data. This integration streamlines the interaction between the language model and the ROS environment, allowing for seamless execution of commands and retrieval of information. It also abstracts the complexity of tool invocation, making it easier to extend ROSA's capabilities.

System prompts \cite{web:openai} (Figure \ref{code:system-prompt}) play a crucial role in this integration by providing the language model with instructions and context. These prompts define the agent's persona, capabilities, limitations, and critical instructions, ensuring that the language model operates within the intended parameters. An example of system prompts is shown in Figure-\ref{code:system-prompt}.

\begin{figure}
    \centering
    \begin{minted}[fontsize=\small, breaklines, frame=single]{python}
system_prompts = [
    ("system", "You are ROSA, an AI agent that uses ROS tools to..."),
    # Additional prompts...
]
    \end{minted}
    \caption{\textbf{Example system prompt following the OpenAI API standard. The list of system prompts can contain any number of prompts to help guide the agents behaviors, responses, etc.}}
    \label{code:system-prompt}
\end{figure}

\begin{figure}
    \centering
\begin{minted}[fontsize=\small, breaklines, frame=single]{python}
@tool
def add_all(numbers: List[float]) -> float:
    """Returns the sum of numbers."""
    return sum(numbers)

@tool
def mean(numbers: List[float]) -> dict:
    """Returns the mean and standard deviation of a list of numbers."""
    return {
        "mean": statistics.mean(numbers),
        "stdev": statistics.stdev(numbers),
    }
\end{minted}
    \caption{\textbf{Examples of custom tools that go beyond interacting with ROS. These tools help avoid hallucinations by giving the Agent a means to run calculations, rather than relying on fallible notions of calculation from the models training data.}}
    \label{code:calculation_tool_example}
\end{figure}

\subsection{Model Selection and Trade-offs}

The choice of the language model significantly impacts ROSA's performance, capabilities, and resource requirements. ROSA currently supports chat models like \textit{GPT-4o}, \textit{Claude 3.5 Sonnet}, and \textit{Llama 3.2}. The selection between API-based and local inference models depends on the specific needs and constraints of the deployment environment. There are two requirements a model must satisfy in order to be used with ROSA:

\begin{enumerate}[leftmargin=*,labelsep=5pt]
    \item It must support \textit{tool calling} \cite{web:langchain_models}.
    \item It must have a context length of no less than \textit{8192} tokens
\end{enumerate}

For API inference, models like \textit{GPT-4o} are recommended due to their advanced reasoning capabilities and lower cost per token. These models offer superior performance in understanding complex queries and generating accurate responses. However, they require internet connectivity and may raise concerns regarding data privacy and dependency on external services. For local inference, models like \textit{Llama 3.1 8B} provide a balance between performance and resource requirements. Running models locally enhances privacy and reduces reliance on external services, which is beneficial in secure or resource-constrained environments. However, local models may require optimization to achieve real-time performance and may have limitations in handling long context windows or complex reasoning tasks.

When selecting a model, trade-offs between context length, performance, resources, and deployment considerations must be evaluated. Larger models typically offer better performance but require more computational resources. Models with longer context windows can handle extended conversations but may consume more memory and processing power. Deployment considerations, such as the need for offline operation or data privacy requirements, may influence the choice of a local model over an API-based one.

\begin{figure}
    \centering
    \begin{minted}[fontsize=\small, breaklines, frame=single]{python}
@tool
def read_log(
    log_file_directory: str,
    log_filename: str,
    level_filter: Optional[str] = None,
    num_lines: Optional[int] = None
) -> dict:
    """
    Reads a log file and returns entries matching the specified criteria.

    :param log_file_directory: directory...
    :param log_filename: name of the...
    :param level_filter: filter logs by...
    :param num_lines: number of lines...
    """
    # Implementation details...
    \end{minted}
    \caption{\textbf{Example function signature for a tool that extends beyond ROS interactions. This particular tool can be used to read log files, allowing ROSA to find errors, warnings, and other commonly logged and informative information.}}
    \label{code:log_tool_example}
\end{figure}

\subsection{Design Choices}

Several key design decisions were made during the implementation of ROSA to enhance usability, safety, and effectiveness. The modular architecture--separating tool packages into \texttt{ROS1}, \texttt{ROS2}, \texttt{calculation} (Figure \ref{code:calculation_tool_example}), \texttt{logging} (Figure \ref{code:log_tool_example}), and other modules--simplifies maintenance and allows for clear separation of version-specific functionalities. This organization facilitates easier updates and scalability as ROS evolves.

By automating aspects like parameter injection and tool registration, the ROSA library reduces the potential for errors and streamlines the development process by design. Parameterization of tools with options like patterns, name spaces, and blacklists provides flexibility while maintaining control over operations. Allowing the language model to specify these parameters enables ROSA to perform precise actions as instructed by the user, enhancing the agent's usefulness in varied scenarios. Returning structured data enhances the language model's ability to generate accurate responses and reduces the risk of confabulation. This design choice is critical in ensuring that operators receive reliable information, thereby increasing trust in the system. 

The integration with LangChain and the ReAct framework abstracts the complexity of tool invocation and reasoning processes, allowing developers to focus on extending ROSA's capabilities rather than managing low-level interactions. This abstraction layer simplifies the agent's architecture and promotes more rapid development and experimentation. Prioritizing safety through parameter validation, constraint enforcement, and centralized safety mechanisms addresses potential risks associated with executing commands in the ROS environment.

%% file: ethical_considerations.tex
The development and deployment of embodied agents, such as ROSA, raise significant ethical considerations. As robots gain autonomy and interact with complex environments, it becomes crucial to define frameworks that promote ethical behavior, protect human welfare, and maintain public trust. This section outlines principles, practices, and open questions that guide the safe and responsible adoption of such technologies, with particular emphasis on Asimov's Three Laws of Robotics.

\subsection{Foundational Ethical Principles and Asimov’s Laws}

Isaac Asimov’s \textit{Three Laws of Robotics} \cite{web:asimov} provide a widely recognized (albeit fictional) starting point for ethical guidelines in robotics:

\begin{enumerate}[label=\textbf{Law \arabic*:}, leftmargin=*, parsep=2pt, itemsep=2pt]
    \item A robot may not injure a human being or, through inaction, allow a human being to come to harm.
    \item A robot must obey orders given by human beings except where such orders would conflict with the First Law.
    \item A robot must protect its own existence as long as such protection does not conflict with the First or Second Law.
\end{enumerate}

Practical implementations of these laws in real-world systems must contend with complexities such as interpreting user intent, balancing trade-offs in uncertain environments, and enforcing directives that may occasionally conflict. Below, we discuss how the ROSA framework addresses---or in some cases does not fully address---each of these laws.

\subsubsection{Addressing the First Law (Preventing Harm to Humans)}

ROSA is primarily an AI-agent framework that coordinates high-level decision-making. It does not, by default, include active ``uplink'' modules for physical control; only ``downlink'' modules for system observation are provided out of the box. Thus, the framework encourages developers to implement their own control pathways, including mechanisms to detect and avoid human injury. Some ways in which ROSA supports the First Law are:
\begin{itemize}[leftmargin=*]
    \item \textbf{Mandatory E-Stop and Overrides:} ROSA-based systems can (and should) be integrated with external e-stop devices or software overrides. These can immediately halt any action, mitigating the possibility of harm due to misinterpretation of data or erroneous commands.
    \item \textbf{Fail-Safe Defaults:} In ambiguous or hazardous conditions, ROSA can revert to a safe or idle state until it receives clarification from a human operator. This behavior aims to reduce the risk of unintended harmful actions.
    \item \textbf{Encouraged Integration with Safety Layers:} Although not enforced within ROSA itself, developers are strongly advised to employ lower-level ROS packages for human detection, collision avoidance, workspace monitoring, and other safety-critical functions.
\end{itemize}
However, ROSA alone cannot guarantee compliance with the First Law unless it is coupled with the appropriate sensing and hardware safety features. Responsibility ultimately lies with developers and operators to integrate these capabilities effectively.

\subsubsection{Addressing the Second Law (Obedience to Humans)}

ROSA supports a hierarchical command structure in which human operators retain ultimate control. By default, any directive from a human user supersedes the AI agent's plan. This can be realized through:
\begin{itemize}[leftmargin=*]
    \item \textbf{Command Multiplexers:} These modules listen to both ROSA-generated commands and human input (e.g., joystick or teleoperation). If a conflict arises, human commands take precedence and ROSA’s commands are overridden.
    \item \textbf{Explainable Decision-Making:} ROSA logs and exposes its decision-making traces, enabling operators to review and revise or halt upcoming actions. This transparency ensures that human oversight remains effective and informed.
\end{itemize}
Despite these measures, it remains possible for humans to issue harmful or contradictory commands. ROSA cannot independently resolve deliberate misuse or ill-intentioned directives. Additional policy-level controls or organizational oversight are advised to address such scenarios.

\subsubsection{Addressing the Third Law (Preserving the System’s Existence)}

ROSA includes a range of self-diagnostic and monitoring tools that enable an agent to detect and report anomalies in the environment or in the robot’s hardware. Within an integrated robotic system, ROSA can:
\begin{itemize}[leftmargin=*]
    \item \textbf{Monitor Critical System Health:} The agent can be programmed to detect sensor failures, battery anomalies, or actuator malfunctions and then engage mitigation behaviors (e.g., moving to a safe state, shutting down power to certain subsystems).
    \item \textbf{Prioritize Self-Preservation Consistent with Human Safety:} In the event of certain high-risk conditions, ROSA can pause all non-critical tasks and focus on preserving the robot’s functional integrity, as long as doing so does not conflict with higher-level safety concerns for humans.
\end{itemize}
This layered approach ensures that the agent can protect itself, but not at the expense of human safety or legitimate human instructions.

\subsection{Safety and Risk Mitigation}

\subsubsection{Human Override and E-Stop}
Human overrides and e-stop features are essential for mitigating accidental harm. ROSA allows for external e-stop signals that immediately suspend or cancel robotic motion. Command multiplexers or hardware interlocks can force all high-level commands to cease. These practices align with the First Law by preventing unwanted or potentially harmful actions.

\subsubsection{System-Specific Safety Considerations}
While ROSA can interface with standard ROS packages for collision avoidance and human-detection, it does not enforce these by default. Each deployment must integrate low-level safety features that match its specific application, such as sensors for proximity alerts, vision-based human tracking, and compliance-based motion controllers. ROSA’s high-level autonomy thus complements rather than replaces existing safety modules.

\subsubsection{Gaps and Future Enhancements}
Although ROSA reduces risk by separating ``downlink'' (observation) from ``uplink'' (actuation) capabilities, there is no absolute technical barrier to misuse if an operator creates modules that issue harmful commands. Future iterations of ROSA may introduce permission layers, code review pipelines, or curated tool repositories that clearly demarcate ``trusted'' from ``untrusted'' functionalities. More rigorous user authentication and code signing could also help mitigate malicious or accidental misuse.

\subsubsection{Open-Source Governance and Ethical Oversight}

ROSA is distributed as open-source software, which fosters collaboration and rapid innovation but also necessitates diligent governance. Currently, repository administrators review pull requests, and submissions that enable harmful behaviors are disallowed. Future plans include an ethics-focused discussion forum, more explicit contributor guidelines, and automated checks for dangerous functionalities. These measures aim to prevent the addition of modules that could violate Asimov’s Laws or general ethical norms.

\subsection{Operability and Need for Extended Case Studies}

ROSA’s present design has been validated in small scale proofs of concept, but more comprehensive multi-robot or long-duration experiments will be necessary to quantify improvements in efficiency, safety, and reliability. Proposed metrics include:
\begin{itemize}[leftmargin=*]
    \item \textbf{Mean Time Between Human Interventions (MTBHI):} A measure of how often operators must manually override the system.
    \item \textbf{Task Completion Rate:} The fraction of tasks successfully completed under normal autonomous conditions.
    \item \textbf{Incident Reporting Rate:} The frequency of safety-related events or near-misses.
\end{itemize}
Such metrics could clarify the impact of the ROSA framework on real-world applications and demonstrate how effectively it upholds Asimov’s principles in practice. Longer-term or field-based case studies will also provide opportunities to refine system components, expand the safety feature set, and address evolving ethical challenges.

In summary, while ROSA embraces the spirit of Asimov’s Laws by prioritizing human safety, respecting user directives, and enabling self-protective features, complete adherence to these laws depends on hardware integration, robust sensing, and diligent governance practices. Ongoing enhancements to framework features, coupled with rigorous field testing, will improve both operability and compliance with these foundational principles.

%% file: conclusion.tex
In this paper, we introduced ROSA, a novel AI agent that bridges the gap between robotic systems and human operators through natural language interfaces. By leveraging advanced language models and open-source frameworks, ROSA enables users to interact with complex robotic platforms using everyday language, thereby enhancing both accessibility and operational efficiency. We described ROSA’s agent architecture, detailing its action space, system prompts, and tool invocation mechanisms. The modular and cross-compatible design supports integration with both ROS1 and ROS2, underscoring its adaptability. Furthermore, structured tools and safety mechanisms ensure that ROSA operates reliably and securely within diverse ROS environments.

ROSA represents a major step forward in human-robot interaction, offering operators immediate access to extensive system knowledge and multimodal communication capabilities, including speech, visual perception, and multilingual usage. By addressing the challenges arising from the complexity and specialization of robotic systems, ROSA empowers users with varying levels of expertise to interact more effectively with robots. We also explored the ethical considerations pertinent to AI agents in robotics, emphasizing the need for safety, transparency, privacy, and responsible usage. By integrating foundational ethical principles and proposing community guidelines, we aim to support the creation of robust ethical frameworks that can guide future innovations in AI and robotics.

Looking ahead, ROSA’s development can be advanced in several ways. Future efforts include improving contextual understanding, refining disambiguation strategies, and strengthening the agent’s ethical decision-making processes. Investigations into fine-tuned small language models may reduce reliance on hosted models, thereby enhancing privacy and expanding offline capabilities, although challenges related to latency, model performance, and the increased hardware costs of local deployment must be addressed. Furthermore, incorporating semantic and episodic memory, as inspired by the Cognitive Language Agent methodologies, could enable ROSA to maintain more comprehensive context, make better-informed decisions, and learn more effectively from its mistakes. By fostering cross-disciplinary collaboration and upholding ethical standards, we envision a future in which embodied agents like ROSA play an integral role in advancing robotics, benefiting society at large, and promoting responsible technological progress.

%% file: biography.tex
\begin{biographywithpic}
{Rob Royce}{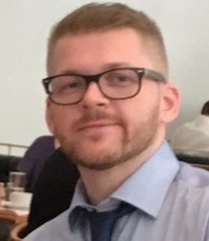} received his degree in Computer Science and Engineering from the University of California, Los Angeles (UCLA), where he was admitted as a Regents Scholar. He is currently a Data Scientist in the Artificial Intelligence \& Analytics Group at NASA's Jet Propulsion Laboratory (JPL) and serves as the Systems Integration Lead for various robotics projects, including NAISR/NeBula and EELS. Rob is a leading practitioner and subject matter expert in Generative AI (GenAI) at JPL, and is also registered as a NASA AI Advisor. He has worked on numerous innovative GenAI projects, such as utilizing retrieval-augmented generation to analyze vast datasets and creating applications that enable users to interact with their documents through conversational interfaces. His extensive experience in software engineering, artificial intelligence, machine learning, and robotics has culminated in the development of ROSA, the Robot Operating System Agent.
\end{biographywithpic}

\begin{biographywithpic}
{Marcel Kaufmann}{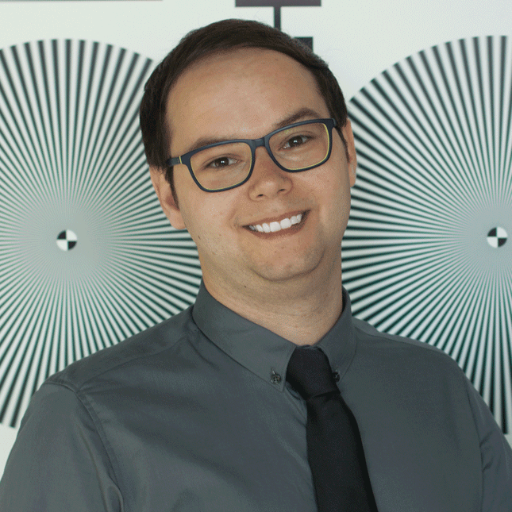} is a Data Scientist in the Future Technology Exploration and Infusion Group at NASA's Jet Propulsion Laboratory (JPL). His research focuses on enabling innovative technologies in multi-robot systems, human-robot interaction, remote sensing, and data visualization. He holds a Ph.D. (2024) in Computer Engineering from Polytechnique Montreal, Canada, and an M.Sc. (2016) and a B.Sc. (2015) in Photonics and Computer Vision from the University of Applied Sciences Darmstadt, Germany. He previously developed an autonomy assistant for multi-robot operations called Copilot MIKE as part of the DARPA SubT Challenge. He is also a former Vanier Canada Graduate Scholar (2019) and an alumnus of the International Space University's Space Studies Program (2017), hosted by the Cork Institute of Technology in Ireland.
\end{biographywithpic}

\begin{biographywithpic}
{Jonathan Becktor}{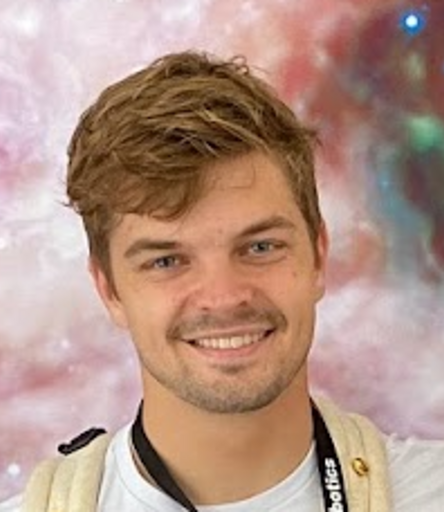} is a Robotics Technologist in the Maritime and Multiagent Systems Group at NASA’s Jet Propulsion Laboratory (JPL). His research centers on developing advanced autonomy and perception systems for multi-agent robotics, with key roles in the NAISR/NeBula program and the DARPA LINC challenge. Jonathan's work focuses on robust object detection, semantic mapping, and integrating machine learning techniques for autonomous systems. He holds a Ph.D. (2023) in Computer Science and Electrical Engineering from the Technical University of Denmark. Prior to JPL, he worked as a machine learning engineer and conducted research in perception technologies for robotics.
\end{biographywithpic}

\begin{biographywithpic}
{Sangwoo Moon}{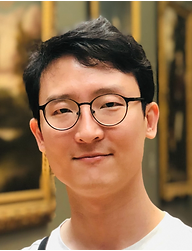} is a Postdoctoral Fellow in the Mobility and Robotic Systems at NASA Jet Propulsion Laboratory (JPL). His primary focus is on developing target belief model representation from multi-modal sensor data and leveraging learning-based decision making for proactive exploration and target localization using vision-language-action models (VLAM). His research interests encompass uncertainty-aware and information-driven robotic mission and path planning, capturing information measure in stochastic environments, and communication-aware informative planning for multi-robot systems. He received a Ph.D. from the Ann and H.J.Smead Aerospace Engineering Sciences at the the University of Colorado Boulder (2021).
\end{biographywithpic}

\begin{biographywithpic}
{Kalind Carpenter}{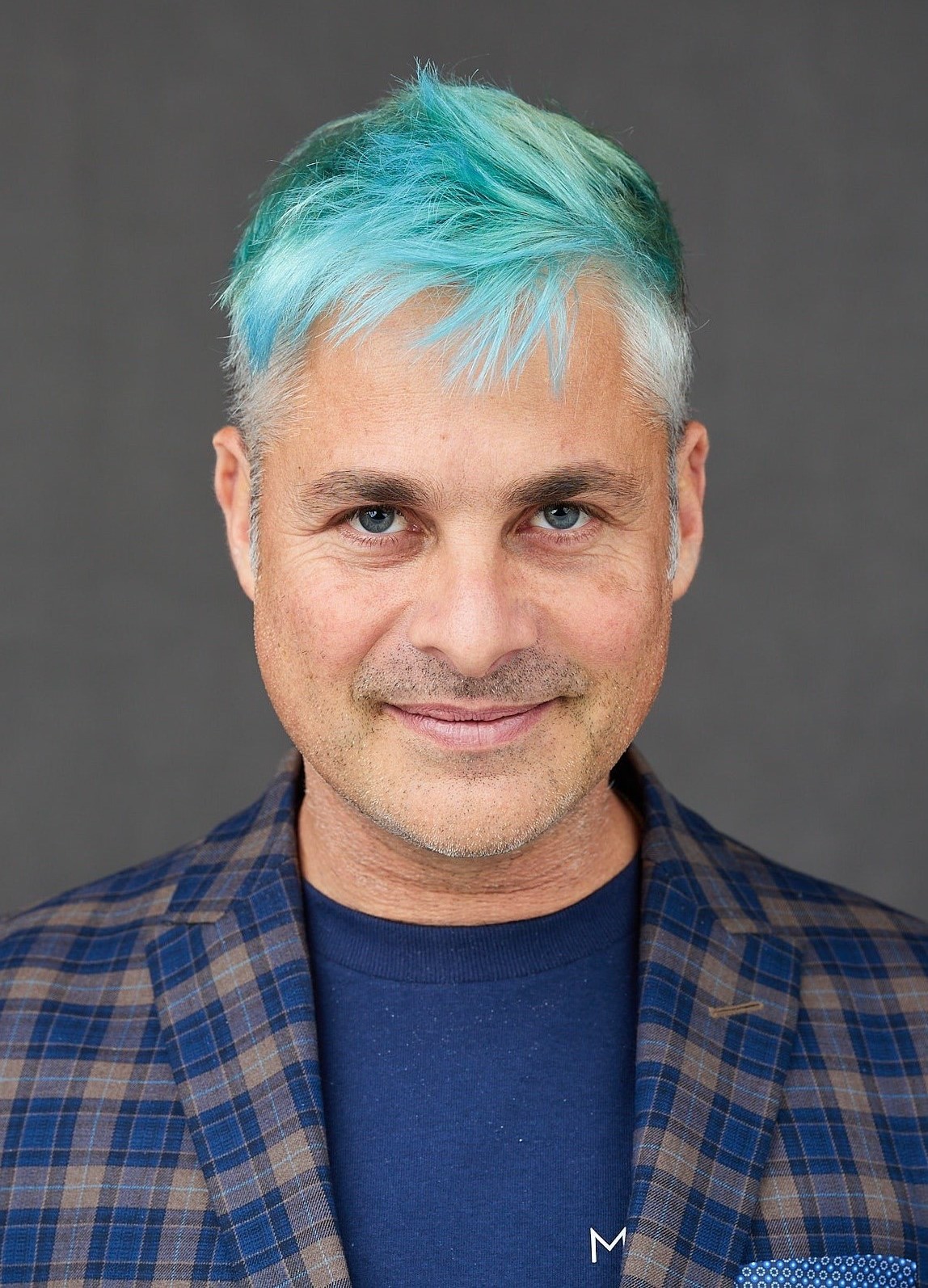}
Kalind is a Robotics Engineer in the robotic Advanced Robotic Systems group at JPL. He focuses on robotic access, manipulation and sampling technologies. His latest work includes the sample tube gripper lead for the Sample Recovery Helicopter. Kalind is the inventor and was the Principal Investigator of the Exobiology Extant Life Surveyor (EELS), an adaptable mobility capability aimed to traverse through the plume vent crevasses on Enceladus to reach the ocean below the ice. Previous work includes co-inventor of PUFFER and principle investigator (PI) of an autonomous swarm of robots with ground penetrating radar to map the changes in sub glacial topology in Greenland and Antarctica created by ocean currents that directly. This work was the precursor to the current CADRE mission.

\end{biographywithpic}

\begin{biographywithpic}
{Kai Pak}{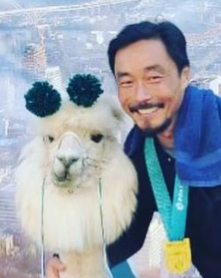} is a Data Scientist in the Artificial Intelligence \& Advanced Analytics Group at NASA's Jet Propulsion Laboratory (JPL) and co-lead for the Lab's exploration and experimentation with Large Language Models for scientific, engineering, and institutional applications. His research focuses on developing computer vision capabilities for satellite and other remote sensing imagery for Earth and planetary science applications. He holds an M.S. in Applied Physics from Columbia University and an M.S. in Computer Science from the University of Illinois Urbana-Champaign (UIUC).  
\end{biographywithpic}

\begin{biographywithpic}
{Amanda Towler}{towler.png} is the Group Supervisor for the Future Technology Exploration and Infusion Group at JPL, where she co-leads the lab's exploration of Large Language Models and GenAI capabilities. Prior to joining JPL, Amanda was an Associate Researcher with the Applied Research Laboratory for Intelligence \& Security (ARLIS). Previously, she was the Co-Founder and Principal Investigator of Hyperion Gray LLC, a woman-owned small business providing technology R\&D, rapid prototyping, and independent security research services to the US Government.
\end{biographywithpic}

\begin{biographywithpic}
{Rohan Thakker}{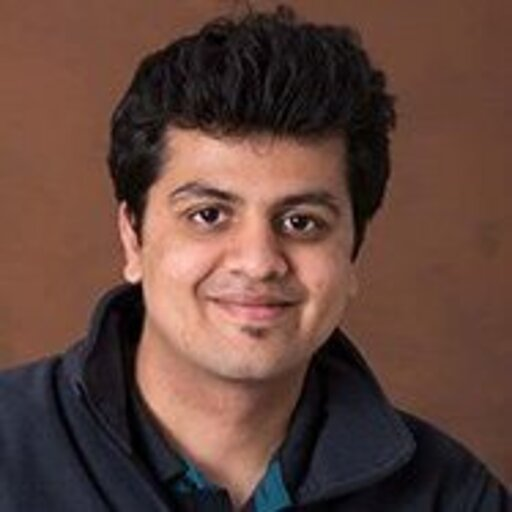} is a Robotics Technologist in the Robotics and Mobility Group at NASA's Jet Propulsion Laboratory since 2017 after finishing his graduate studies in robotics at Carnegie Mellon University. 
His research focuses on autonomous decision-making under uncertainty in real-world environments across the entire robotics stack.
He is the Autonomy Lead of the EELS Snake Robot team which developed and deployed the NEO autonomy stack for exploring unknown ice worlds at the Athabasca Glacier in Canada. 
He has led autonomy sub-system teams across three DARPA programs including team costar that finished first in DARPA Subterranean Challenge's Tunnel Circuit. 
\end{biographywithpic}

\begin{biographywithpic}
{Shehryar Khattak}{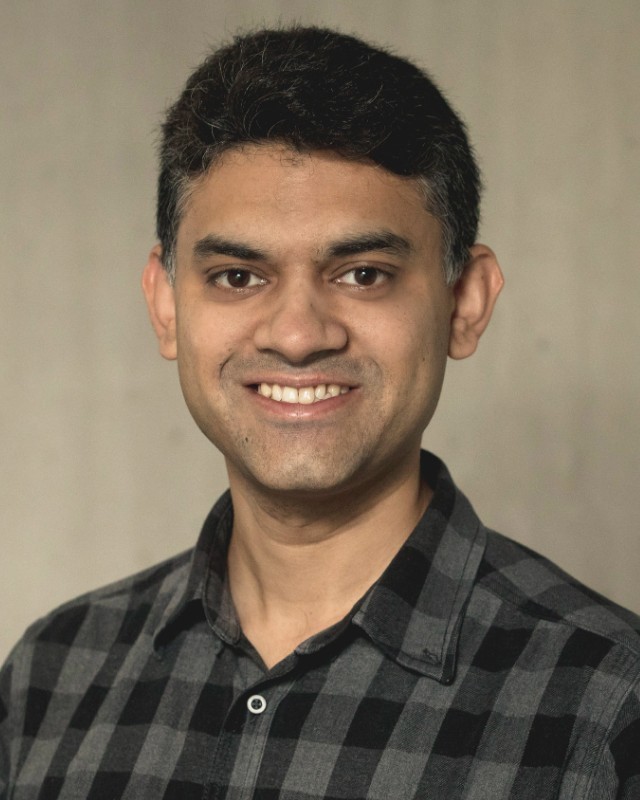} is a Robotics Technologist in the Perception Systems Group at NASA's Jet Propulsion Laboratory (JPL). His work focuses on enabling resilient robot autonomy in complex environments through multi-sensor information fusion. Currently, he is the Principal Investigator (PI) for the Multi-robot Autonomous Intelligent Search and Rescue task at JPL. He previously served as the perception lead for JPL's team in the DARPA RACER project and for Team CERBERUS, winners of the DARPA Subterranean Challenge. Before joining JPL, Shehryar was a postdoctoral researcher at ETH Zurich. He earned his Ph.D. (2019) and M.S. (2017) in Computer Science from the University of Nevada, Reno Additionally, he holds an M.S. in Aerospace Engineering from KAIST (2012) and a B.S. in Mechanical Engineering from GIKI (2009).
\end{biographywithpic}